\documentclass[conference]{IEEEtran}
\IEEEoverridecommandlockouts
\usepackage{cite}
\usepackage{amsmath,amssymb,amsfonts}
\usepackage{algorithmic}
\usepackage{graphicx}
\usepackage{textcomp}
\usepackage{multicol}
\usepackage{multirow}
\usepackage{soul}
\usepackage{pifont}
\usepackage{bbding}

\usepackage[table]{xcolor}
\usepackage{booktabs}
\usepackage{authblk}
\usepackage[hidelinks, bookmarks=false]{hyperref}

\renewcommand{\footnoterule}{%
  \kern -3pt
  \hrule width 0.2\textwidth height 0.5pt
  \kern 2pt
} 

\def\BibTeX{{\rm B\kern-.05em{\sc i\kern-.025em b}\kern-.08em
    T\kern-.1667em\lower.7ex\hbox{E}\kern-.125emX}}
\begin{document}

\title{CCUP: A Controllable Synthetic Data Generation Pipeline for Pretraining Cloth-Changing Person Re-Identification Models}


\author {
    Yujian Zhao\textsuperscript{1}, Chengru Wu\textsuperscript{2}, 
    Yinong Xu\textsuperscript{2}, Xuanzheng Du\textsuperscript{2},
    Ruiyu Li\textsuperscript{2}, Guanglin Niu\textsuperscript{1*}
    \\
    \textsuperscript{1} School of Artificial Intelligence, Beihang University \\
    \textsuperscript{2} Shen Yuan Honors College, Beihang University \\ 

    \{yjzhao1019, chengru\_wu, yinong\_xu, xuanzheng\_du, ruiyu\_li, beihangngl\}@buaa.edu.cn 
    \thanks{\textsuperscript{*}Correspinding author. This work was supported by the National Natural Science Foundation of China (No. 62376016).}
}

\maketitle

\renewcommand{\thefootnote}{}

\begin{abstract}
Due to the high cost of constructing Cloth-changing person reidentification (CC-ReID) data, the existing data-driven models are hard to train efficiently on limited data, which causes the issue of overfitting. To address this challenge, we propose a low-cost and efficient pipeline specific to CC-ReID tasks for generating controllable and high-quality synthetic data simulating the surveillance scenarios. Particularly, we construct a new self-annotated CC-ReID dataset named Cloth-Changing Unreal Person (CCUP), containing 6,000 IDs, 1,179,976 images, 100 cameras, and 26.5 outfits per individual. Based on this large-scale dataset, we introduce an effective and scalable pretrain-finetune framework for enhancing the generalization of the traditional CC-ReID models. The extensive experimental results demonstrate that our framework could improve the original models such as two typical models TransReID and FIRe$^2$ after pretraining on CCUP and finetuning on a benchmark, and outperform other state-of-the-art models. The dataset is available at: \href{https://github.com/yjzhao1019/CCUP}{https://github.com/yjzhao1019/CCUP}.
\end{abstract}


\section{Introduction}
\label{sec:intro}

Person re-identification (ReID) aims to identify gallery images containing persons with the same identity as the query image in a cross-camera scenario. Furthermore, cloth-changing person re-identification (CC-ReID) is a more challenging task to identify the same person but with different clothes in real-world scenarios at large spatial and temporal scales.

\begin{table*}[htbp]
\label{table1}
\caption{Statistic of CC-ReID and synthetic datasets. Hyphens represent the number of outfits is not provided.}
\begin{center}
\begin{tabular}{ccccccc}
\toprule


Characteristic & Dataset  & \#IDs    & \#Images  & \#Cam & \parbox{2cm}{\centering Surveillance \\ Simulate} & \#avgClo \\
\midrule
\multirow{9}{*}{Real for CC-ReID} &PRCC \cite{PRCC}  & 221    & 33,698  & 3 & \checkmark  &     2.00   \\ 
& Celeb-reID \cite{Celeb-reID}       & 1,052  & 34,186  & -   &\ding{55} &      -  \\
& Celeb-reID-light \cite{Celeb-reID}  & 590    & 10,842  & -  &\ding{55} &      - \\ 
& LTCC \cite{LTCC}              & 152    & 17,138  & 12  & \checkmark &      3.14  \\

& DeepChange \cite{DeepChange}        & 1,124  & 178,407 & 17 & \checkmark &      2.62  \\ 
& LaST \cite{LaST}              & 10,862 & 228,000 & -   &  \ding{55} &   -  \\
& NKUP \cite{NKUP}             & 107    & 9,738   & 15  &  \checkmark &    -  \\
& NKUP+ \cite{NKUP+}       & 361    & 40,217  & 29  &   \checkmark & -     \\
\midrule
\multirow{4}{*}{Synthetic for ReID} 
& PersonX \cite{PersonX}          &1,266   &273,456  & 6    & \ding{55} &      - \\
& RandPerson \cite{RandPerson}        & 8,000  &1,801,816  &19   &\checkmark &      - \\
& UnrealPerson \cite{UnrealPerson}     & 6,799  &1,256,381&34   & \checkmark &     -  \\
& ClonedPerson \cite{ClonedPerson}     & 5,621  &887,766  &24   & \checkmark &     -  \\
\midrule
\multirow{2}{*}{ Synthetic for CC-ReID} 
& VC-Clothes \cite{VC-Clothes}        & 512    & 19.060  & 4   & \checkmark &      2.07  \\
& \cellcolor{green!20} CCUP (ours)   & \cellcolor{green!20} 6000   & \cellcolor{green!20} 1,179,976       &  \cellcolor{green!20}100  & \cellcolor{green!20} \checkmark &  \cellcolor{green!20} \textbf{26.5}\\
\bottomrule
\end{tabular}
\end{center}
\end{table*}

In recent years, deep learning-based models \cite{ResNet, Transformer, ViT} have been widely used to learn the discriminative features of person images for ReID and its extended task CC-ReID \cite{TransReID} . However, there are two main challenges for the CC-ReID task. 

\textbf{Challenge 1: the high cost of sampling and labeling real CC-ReID images limits the size of existing datasets, causing low performance of training models due to the lack of sufficient ground truth for supervision.} Building a ReID dataset requires a complex environmental setup of places, devices and pedestrians, as well as manual labeling without violating privacy (DukeMTMC-ReID has been retracted due to privacy concerns). In particular, the complexity and costs of generating a CC-ReID dataset further increase significantly since it is difficult to capture images of the same person wearing various outfits on a large spatial and temporal scale. In contrast, synthetic datasets are emerging to reduce costs and address privacy concerns. As shown in Tab.~\ref{table1}, we provide the statistic of identities (\#IDs), images (\#Images), cameras (\#Cam) and average outfits per identity (\#avgClo) of some typical CC-ReID and synthetic datasets. We could observe that the whole size and especially \#avgClo of all the previous CC-ReID benchmark datasets such as PRCC, LTCC and VC-Clothes are obviously limited, and Celeb-reID and LaST are even created from celebrity street photography and movies rather than real surveillance scenes. Besides, few existing commonly-used synthetic datasets are not designed for cloth-changing scenarios and therefore lack rich cloth-changing ground truth. \textbf{To address these issues, we propose a controllable and low-cost pipeline for generating large-scale synthetic data more suitable for CC-ReID tasks.}

\textbf{Challenge 2: cloth-irrelevant features are hard to be extracted via the existing models straightly trained on a limited CC-ReID dataset.} Specific to the CC-ReID task, the most pivotal purpose is to extract discriminative cloth-irrelevant features. Therefore, CAL\cite{CAL} is proposed to extract cloth-irrelevant features from original RGB images by penalizing the predictive power of the ReID model. AIM \cite{AIM} is proposed to analyze the impact of clothing on model inference and eliminate clothing bias during training . Besides, various auxiliary information such as gait \cite{gait}, skeleton \cite{TranSG}, and 3D shape \cite{3DInvar} could be exploited for supplementing more cloth-irrelevant features. However, all the previous CC-ReID models suffer from extremely scarce training data, limiting their performance specifically some advanced visual transformer-based models. \textbf{To address this challenge, we employ a scalable pretrain-finetune framework leveraging our large-scale synthetic dataset to enhance the model performance of CC-ReID.}

Overall, the contributions of our work are three-fold: 
\begin{itemize}
    \item We construct a high-quality synthetic CC-ReID dataset named CCUP with our low-cost and controllable data generation pipeline, which is the first large-scale (over 1,000,000 images) dataset for the CC-ReID task. 
    \item We exploit a scalable pretrain-finetune framework, which could improve the performance of CC-ReID via fine-tuning the same model pretrained on our large-scale synthetic dataset CCUP.
    \item The extensive experimental results on multiple benchmark datasets including LTCC, VC-Clothes and NKUP illustrate that our framework outperforms other state-of-the-art baseline models significantly and consistently.
\end{itemize}

\section{Related Work}
\subsection{Cloth-changing person re-identification}

Traditional ReID studies are highly dependent on clothing appearance, which is not available to address unstrained clothing changes in real scenarios. Thus, there are an increasing number of researches related to Cloth-Changing person Re-Identification (CC-ReID). Researchers first recognize the importance of data for the task and therefore construct many benchmark datasets \cite{PRCC, Celeb-reID, LTCC, DeepChange, LaST, NKUP, NKUP+, VC-Clothes}. Then, many innovative approaches have been proposed, the core idea of which is to focus on cloth-irrelevant features. TransReID \cite{TransReID} propose a pure transformer-based object ReID framework containing novel modules such as jigsaw patch module and side information embeddings. FSAM \cite{FSAM} propose a two-stream framework that learns discriminative body shape knowledge and transfers it to complement the cloth-unrelated knowledge. Pos-Net \cite{Pos-Neg} reinforces the feature learning process by designing powerful complementary data augmentation strategies. IGCL~\cite{IGCL} proposed a novel framework where the human semantics are leveraged and the identity is unchangeable to guide collaborative learning. IRM~\cite{IRM} propose a new instruct-ReID task and a large-scale OmniReID benchmark as well as adaptive triplet loss. Pixel sampling~\cite{pixel} propose a semantic-guided approach that forces the model to automatically learn cloth-irrelevant signals by randomly changing clothes pixels. IS-GAN~\cite{IS-GAN} disentangles identity-related and unrelated features from person images through an identity-shuffling technique that exploits identification labels. FIRe$^2$ \cite{FIRe2} designs a fine-grained feature mining module and presents a fine-grained attribute recomposition module by recomposing image features with different attributes. IFD \cite{IFD} proposes an Identity-aware Feature Decoupling learning framework to mine identity-related features. However, these models have never been able to identify cloth-changing person well due to the limited data.

\subsection{Dataset synthesis}

Techniques for dataset synthesis in CC-ReID domain are mainly categorized into traditional graphics methods and deep learning methods. Traditional graphics methods first generate 3D meshes, which are then imported into a physics engine to configure animations, add scenes, and simulate real surveillance \cite{PersonX, RandPerson, UnrealPerson, ClonedPerson, VC-Clothes}. However, these methods still have some shortcomings due to the inability of existing software to generate high-resolution meshes on large scales. Deep learning methods commonly use GAN \cite{GAN} for continuous iteration to synthesize dataset. CCPG \cite{CCPG} proposes a GAN based model for clothing and pose transfer across identities to augment images of more clothing variations. AFD-Net \cite{AFD-Net} proposes a novel framework containing intra-class reconstruction and inter-class adversary to disentangle the identity-related and identity-unrelated features. What's more, although diffusion model \cite{diffusion} has received a lot of attention in the field of image synthesis in recent years, there is no more mature work for CC-ReID.


\begin{figure*}[htbp]
\centerline{\includegraphics[width=1.0\textwidth]{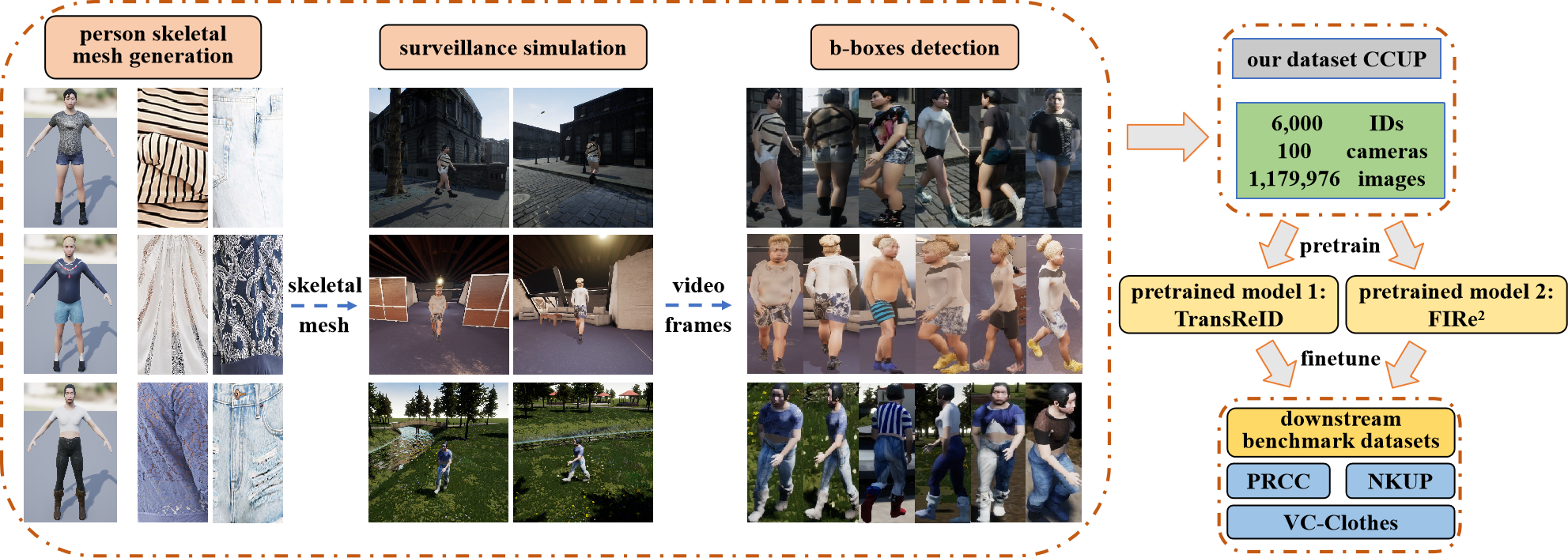}}
\caption{Pipeline of our work. We first generate the skeletal mesh of person and provide a large number of clothing textures for cloth-changing. Then the skeletal mesh is imported into the three scenarios to simulate surveillance and the person in the surveillance frames is detected. Finally, we construct a large-scale dataset called CCUP containing 6,000 IDs, 100 cameras, and 1,1179,976 images. We pretrain the TransReID and FIRe$^2$ two baseline models with CCUP and finetune on PRCC, VC-Clothes, and NKUP benchmarks.}
\label{framework}
\end{figure*}

\section{Methodology}
\label{methodology}

\subsection{Dataset generation}

Considering the numerous advantages of synthetic data, such as good controllability, a high degree of automation, significantly lower costs compared to capturing real-world data, and the excellent ability to simulate various real-world environments, we prioritize synthetic data generation for CC-ReID tasks. Accordingly, we propose a low-cost, high-quality, large-scale, and controllable data generation pipeline along with a new dataset CCUP for pretraining CC-ReID models. Specifically, the CCUP generation process consists of three main procedures: (1) generating skeletal meshes of realistic human characters, (2) simulating surveillance in diverse scenarios, and (3) producing self-labeled detection results for cloth-changing pedestrian, as illustrated in Fig~\ref{framework}.

\subsubsection{Generate person skeletal mesh}
\label{section 2.1}

The skeletal mesh is the basis for building synthetic data, which contains mesh data and skeletal data. Mesh data stores vertex positions, normal vectors and texture coordinates, etc. Skeletal data represents the skeletal node hierarchy. Specifically, we construct the person skeletal mesh using MakeHuman\footnote{https://github.com/makehumancommunity} software, an open-source 3D human modeling software that helps users create high-quality, vividly realistic human body models. 

We modify Makehuman's AssetDownloader plugin and employ it to collect assets from the MakeHuman community that can be used to build 3D person models such as skin, hair, clothes, etc. Besides, we modify MakeHuman's MassProduce plugin to create 6000 naked person skeletal meshes with different physiological features. In this way, unique combinations of physiological parameters determine unique person IDs, so we consider each naked person skeletal mesh to be an ID and we get the set of IDs for the dataset $ID = \{id_1, id_2, ..., id_n\}$, where $n$ is the number of identities in the dataset. Benefiting from our modification of the AssetDownloader plugin, we collecte almost 3,000 clothing asset and get different ensemble of outfits $Clo = \{clo_1, clo_2, ..., clo_m\}$, where $m$ denotes the number of ensemble of outfits. In turn, we can construct clothed skeleton meshes $CSM = \{csm_{11}, csm_{12}, ..., csm_{1t_1}, ..., csm_{n1}, csm_{n2}, ..., csm_{nt_n}\}$:
\begin{equation}
csm_{ij}= \{DressUp(id_i, clo_j) \mid 1 \le i \le n, 1 \le j \le t_i\},
\end{equation}

\noindent where $t_i$ ($t_i \le m$) is the number of clothes for person $id_i$ and $DressUp(id_i, clo_j)$ denotes wearing clothes $clo_j$ on  person $id_i$. Inspired by Unrealperson \cite{UnrealPerson} , we create more than 10,000 clothes textures, constructing the set of textures for the dataset $T = \{t_1, t_2, ... , t_r \}$ for subsequent clothing changes, where $r$ denotes the number of textures of the dataset. 


\subsubsection{Simulate surveillance in scenarios}
\label{section 2.2}

Unreal Engine\footnote{https://www.unrealengine.com/} is a game engine that offers a wide range of rendering functions and is popular in many applications such as game and movie development. To generate the CC-ReID data, we employ Unreal Engine (version 5.3.2) to simulate real surveillance scenarios. We configure animations for the skeletal meshes generated in section \ref{section 2.1} so that they can walk around the simulation scenarios. Then we replace the texture $t_k$ of clothed skeleton meshed $csm$ when they pass by different cameras to simulate the more diverse cloth-changing, denoted as $RT(csm, t)$. Then, we choose three scenarios for simulation in the epic games marketplace: an European alleyway, an office building, and a park with 50, 25, and 25 cameras, respectively. These set of three scenarios are denoted as $S = \{s_1, s_2, s_3\}$.

Particularly, we design travel routes of persons in three scenarios and place cameras along the routes with diverse viewpoints. Benefiting from a well-designed detection strategy, the video of the person could be automatically captured if this person passes under the camera. Then, the set of original frames with automatically labeled person IDs, camera IDs and cloth IDs are generated, denoted as $OF = \{of_1, ..., of_p\}$:
\begin{equation}
    of = Sim(RT(csm, t), s), csm \in CSM, t \in T, s \in S,
\end{equation}

\noindent where $Sim(RT(csm, t), s)$ denotes simulating the real surveillance in scenarios $s$ and obtaining the original frame for $RT(csm,t)$ of surveillance videos. To guarantee the quality of labeled data, each image contains only one pedestrian by adapting the starting time and the speed of this person.

\subsubsection{Detect and label the bounding boxes}

Based on the surveillance video frames $OF$, we employ the advanced RTMdet \cite{RTMdet} model to detect pedestrians and generate their corresponding bounding boxes:
\begin{equation}
    Det(of) =
    \begin{cases}
    (x, y, h, w) & \text{if a person is in } of, \\
    0 & \text{if no person is in } of,
    \end{cases}
\end{equation}

\noindent where $Det(of)$ denotes detecting the person bounding boxes of frames and return the coordinate information if a person is in frames. As explained in section \ref{section 2.2}, each frame contains only one individual, ensuring that the number of bounding boxes per frame is either 0 or 1. Therefore, the frame's label corresponds directly to the label of the bounding box, which significantly simplifies the dataset labeling process. Finally we get our dataset named CCUP $D = \{(I_1, l_1), (I_2, l_2), ..., (I_N, l_N)\}$:
\begin{equation}
    I = of[x:x+w, y:y+h], \text{if } Det(of) \ne 0,
\end{equation}

\noindent where $I$ denotes image of $D$ and $l$ is the ground truth label of the identity.

In our pipeline, we not only change outfits at the mesh body level but also apply diverse texture replacements for the clothes, greatly enhancing the variety of outfits. Besides, our approach is highly extensible, requiring only minor code modifications to generate entirely new datasets tailored to different tasks. The cost of dataset generation is remarkably low due to the high level of automation in the procedure of generating video frames and self-labeling, allowing produce datasets with hundreds of thousands of images in just a few days. Ultimately, we generate a large-scale CC-ReID dataset CCUP, which includes 6,000 individuals, 100 cameras, 1,179,976 images, and an average of 26.5 outfits per individual.

\subsection{Pretrain and finetune}

Pretraining and finetuning are originated in the field of nature language processing, and have gradually been introduced in the field of computer vision, but are rarely used in CC-ReID due to limitation of training data. After obtaining a large amount of CC-ReID data, we introduce the pretrain-finetune framework for CC-ReID tasks. 

The model parameters $\theta$ is first pretrained on a large dataset $D_{pre} = \{(x_1, y_1), ..., (x_p, y_p)\} $, which helps the model capture generalizable features from a broad distribution of data:
\begin{equation}
    \theta^{t+1} = \theta^t - \eta \nabla \mathcal{L}(f_{\theta^t}(x_i), y_i),
\end{equation}
\noindent where $f_{\theta^t}(x_i)$ is model's output for input $x_i$  , $\mathcal{L}$ is the loss function such as cross-entropy or triplet and $\eta$ represents learning rate. After pretraining, we finetune the model on the downstream benchmark dataset $D_{fine} = \{ (\tilde{x_1}, \tilde{y_1}), ..., (\tilde{x_q}, \tilde{y_q})\} $, allows the model to adapt these learned features to the specific task and obtain the best performance parameters $\theta^*$:
\begin{equation}
    \theta^* = \arg \max_{\theta} \mathcal{R}(f_{\theta}(\tilde{x}), \tilde{y}), (\tilde{x}, \tilde{y}) \in D_{fine\_test},
\end{equation}

\noindent where $R$ denotes $Rank$-1 evaluation and $D_{fine\_test}$ is the test set of $D_{fine}$.

\section{Experiment}

\label{experiment}

 We pretrain two models on our CCUP and finetune in the downstream benchmarks to extract more robust cloth-irrelevant features. Furthermore, we demonstrate the superiority of our proposed dataset by comparing the performance of the model pretrained on our CCUP and other synthetic datasets as well as several state-of-the-art baselines.

\subsection{Dataset and evaluation metrics}

For a comprehensive comparison, we select some typical synthetic datasets for pretraining, including UnrealPerson \cite{UnrealPerson}, PersonX \cite{PersonX}, and ClonedPerson \cite{ClonedPerson}. Besides, PRCC \cite{PRCC}, VC-Clothes \cite{VC-Clothes}, and NKUP \cite{NKUP} are chosen as the downstream benchmark datasets for finetuning and obtaining the evaluation results. It is noteworthy that only clothes changing ground truth samples are used in PRCC and VC-Clothes datasets while both clothes-changing and clothes-consistent ground truth samples are used in NKUP. The detailed statistics and comparison of these datasets are shown in Tab.~\ref{table1}. Subsequently, two frequently-used metrics Rank-1 and mAP are employed to evaluate each model.

\subsection{Implementation details}      

We select two representative baselines: a general ReID model TransReID \cite{TransReID} with the backbone ViT \cite{ViT} and a CC-ReID model FIRe$^2$ \cite{FIRe2} with the backbone ResNet50 pretrained on ImageNet \cite{imagenet} as our pretrained models. Besides, the other 11 typical CC-ReID models are utilized for comparison. Particularly, we fix some parameters the same as other baselines for fair comparison. For TransReID, input images are resized to 256 $\times $ 128, the patch size is set to 16, and the stride size is 12. The SGD optimizer is employed with the weight decay of $1 \times 10 ^ {-4}$ and the learning rate is initialized as $4 \times 10^{-3}$. For FIRe$^2$, Input images are resized to 384 $\times$ 192 and the batch size is 32. The Adam optimizer is employed with the weight decay of $5 \times 10 ^ {-4}$ and the learning rate is initialized as $3.5 \times 10^{-4}$. We pretrain the two pretrained models for 10 epochs on UnrealPerson, Personx, ClonedPerson, and CCUP, respectively, and finetune these models on the benchmark datasets for 80 epochs.

\subsection{Comparison with state-of-the-art methods}

\begin{table*}[htbp]
\caption{Comparison of state-of-the-art methods on CC-ReID benchmarks. Bolded values represent the best values obtained on the same baseline. Underline values indicate the overall best performance in each column.}
\begin{center}
\renewcommand\arraystretch{1.2}
\begin{tabular}{c|c|c|cc|cc|cc}
\hline
\multirow{3}{*}{Methods} & \multirow{3}{*}{Backbone}  &\multirow{3}{*}{Pretrain dataset} &      \multicolumn{6}{c} {Datasets}    \\ 
\cline{4-9}
 & &  & \multicolumn{2}{c|}{PRCC (CC) } & \multicolumn{2}{c|}{VC-Clothes (CC) } & \multicolumn{2}{c}{NKUP} \\
\cline{4-9}
 & &  & rank-1 & mAP & rank-1 & mAP & rank-1 & mAP   \\
 
\hline

\hline
PCB (ECCV18) \cite{PCB} & ResNet50  & - & 41.8 & 31.7 & 62.0 & 62.3 & 16.9 & 12.4 \\
MGN (MM18) \cite{MGN} & ResNet50  & - & 25.9 & - & - & - & 18.8 & 15.0 \\
FSAM (CVPR21) \cite{FSAM} & ResNet50  & - & 54.5 & -& 78.6 & 78.9 &- &-  \\
3DSL (CVPR21) \cite{3DSL} & ResNet50  & - & 51.3 & - & 79.9 & 81.2& -&- \\
LSD (IVC21) \cite{LSD} & ResNet50   & - & 37.2 & 47.6 & - & - & 16.4 & 10.2  \\ 
CAL (CVPR22) \cite{CAL} &  ResNet50  & - & 55.2 & 55.8 & 81.4 & 81.7 & - &  - \\

GI-ReID (CVPR22) \cite{GI-ReID} & ResNet50  &- & - & 37.6 & 63.7& 59.0 & -&- \\
AD-ViT (AVSS22) \cite{AD-ViT} & ViT  & - & - & - & - & - & 27.0 & 18.9 \\
AIM (CVPR23) \cite{AIM} & ResNet50  & - & 57.9 & 58.3 & 74.1 & 73.7 & -&  -\\
CCFA (CVPR23) \cite{CCFA} & ResNet50  & - & 61.2 & 58.4 &- &- &- & - \\
IRM (CVPR24) \cite{IRM} & ViT  & - & 54.2 & 52.3 & - & - & -&- \\

\hline
TransReID (ICCV21) \cite{TransReID} & ViT  & - & 45.9 & 48.2 & 73.1 & 74.3 & 21.2 & 14.8  \\
TransReID (ICCV21) \cite{TransReID} & ViT  & UnrealPerson & 51.1 & 50.4 & 72.2  &  74.4 & 26.7 & \underline{\textbf{20.3}}\\
TransReID (ICCV21) \cite{TransReID} & ViT  & PersonX & 46.8 & 47.5 &  72.0 & 73.8 & 25.8 & 19.3 \\
TransReID (ICCV21) \cite{TransReID} & ViT  & ClonedPerson & 47.7 & 48.9 & 73.6 & 71.8 &  27.9 & 19.7  \\
\rowcolor{green!20} TransReID (ICCV21) \cite{TransReID} & ViT  & CCUP (ours) & \textbf{58.9} & \underline{\textbf{59.0}} & \textbf{83.3} & \textbf{83.1}  & \underline{\textbf{28.8}} &  19.4  \\

\hline

FIRe$^2$ (TIFS24) \cite{FIRe2} & ResNet50  & - & 59.1 & 50.5 & 78.0 & 78.9 & 22.4 & 16.0  \\
FIRe$^2$ (TIFS24) \cite{FIRe2} & ResNet50  & UnrealPerson & 62.5 & 57.4 & 84.0 & 84.8 & 24.2 & 16.4  \\
FIRe$^2$ (TIFS24) \cite{FIRe2} & ResNet50  & PersonX & 60.5 & 56.0 & 76.1 & 77.2 & 23.0 & 15.8  \\
FIRe$^2$ (TIFS24) \cite{FIRe2} & ResNet50  & ClonedPerson & 55.9 & 56.4 & \underline{\textbf{85.1}} & \underline{\textbf{85.0}} & 21.8 & 16.0 \\
\rowcolor{green!20}  FIRe$^2$ (TIFS24) \cite{FIRe2} & ResNet50  & CCUP (ours) & \underline{\textbf{64.7}} & \textbf{57.9} & 81.6 & 81.3 & \textbf{26.4} & \textbf{17.8}  \\

\hline
\end{tabular}
\label{table2}
\end{center}
\end{table*}

We compare our model with some state-of-the-art methods in Tab.~\ref{table2}, where CAL, AIM, and CCFA uses person cloth-labels in training process. We observe that both the pretrained models TransReID and FIRe$^2$ empowered by our CCUP illustrate the superior performance than the other models. Specifically, TransReID+CCUP achieves \textbf{59.0\%} mAP on PRCC and \textbf{28.8}\% rank-1 on NKUP, and FIRe$^2$ achieves \textbf{64.7\%} rank-1 on PRCC, and \textbf{85.1\%} rank-1 and \textbf{85.0\%} mAP on VC-Clothes. The best performance of mAP on NKUP is obtained by TransReID pretrained by UnrealPerson, showing the effectiveness and scalability of our pretrain-fine framework. In particular, FIRe$^2$ without pretraining outperforms TransReID, while the performance of TransReID pretrained on our CCPU dataset can be significantly improved. The experimental results demonstrate that our proposed dataset generation scheme and the pretrain-finetune framework facilitate both the general and CC-ReID models to extract more robust and distinguishing identity-relevant features on CC-ReID tasks.

\subsection{Impact of different pretraining dataset}
From the results specific to different synthetic datasets for pretraining as shown in Tab.~\ref{table2}, it is obviously that the performance of TransReID and FIRe$^2$ could be consistently and significantly improved after pretraining on all the datasets. Besides, the model pretrained by CCUP outperforms that pretrained on the other synthetic datasets. Specifically, we exhibit the absolute performance and (improvement over the model without pretraining) of our model. TransReID pretrained by CCUP achieves the best performance with 58.9\% (\textbf{+13.0\%}) rank-1 and 59.0\% (\textbf{+10.8\%}) mAP on PRCC, 83.3\%   (\textbf{+10.2\%}) rank-1 and 83.1\% (\textbf{+8.8\%}) mAP on VC-Clothes, as well as 28.8\% (\textbf{+7.6\%}) rank-1 on NKUP. Furthermore, FIRe$^2$ pretrained by CCUP obtains the best preformance on PRCC and NKUP, achieving 64.7\% (\textbf{+5.6\%}) rank-1, 57.7 (\textbf{+7.2\%}) mAP on PRCC, 26.4\% (\textbf{+4.0\%}) rank-1 and 17.8 (\textbf{+1.8\%}) mAP on NKUP. These results illustrate that the powerful person clothes-changing number of CCUP benefits to performance improvements of Re-ID models.

\subsection{Visualization}

\begin{figure}[htbp]
\centerline{\includegraphics[width=0.5\textwidth]{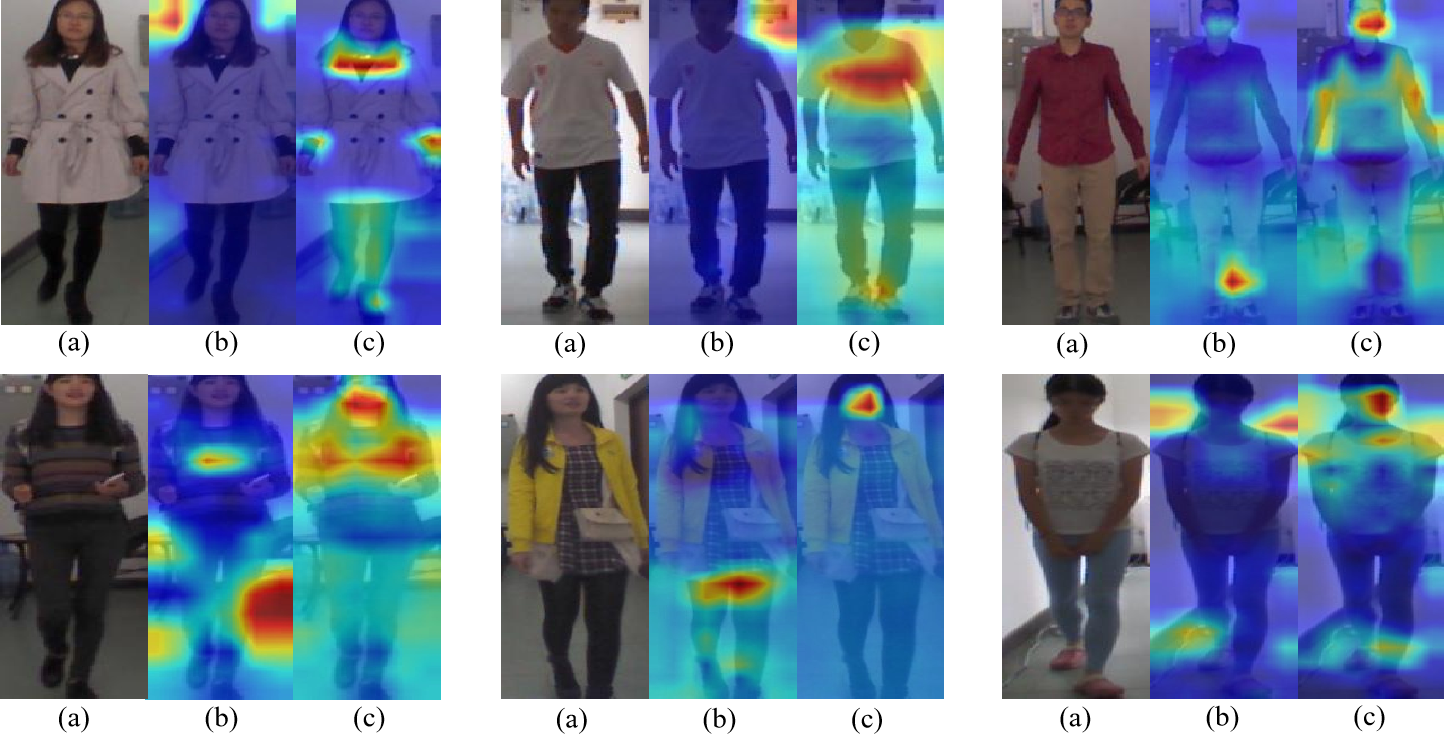}}
\caption{Visualization of FIRe$^2$ evaluated on PRCC. (a) Original iamges. (b) Results without pretraining. (c) Results with pretrain-finetune.}
\label{grad-cam-vis}
\end{figure}

To visually compare the regions that a model fucuses on before and after pre-training, we visualize the inference results of FIRe$^2$ on PRCC as shown in Fig.~\ref{grad-cam-vis} following~\cite{grad-cam}. We observe that the model is improved via pretrained on CCUP in the following two aspects: (1) The model puts more focus on the human body, rather than the background, as illustrated by the three sets of images in the first row of Fig.~\ref{grad-cam-vis}. (2) The model pays more attention to the face, neck, shoulders, wrists, and shoes (Shoes change less frequently than other clothes ) of a person, which are more conducive to extracting cloth-irrelevant features.

\section{Conclusions}
\label{conclusion}

In this paper, we propose a controllable synthetic dataset generation pipeline and construct a CC-ReID dataset named CCUP containing 6000 IDs, 100 cameras, 1,179,976 images, and 26.5 outfits per individual. To overcome the difficulty of extracting robust and accurate cloth-irrelevant features, we introduce the pretrain-finetune framework from NLP. After pretraining both FIRe$^2$ and TransReID baselines, our model outperforms other state-of-the-art methods on three benchmarks: PRCC, VC-Clothes, and NKUP. Furthermore, we conduct comparison experiments using different synthetic datasets as pretrain datasets to demonstrate the superiority of CCUP.

\bibliographystyle{IEEEbib}
\bibliography{icme2025references}

\end{document}